# Review on Application of Drone in Spraying Pesticides and Fertilizers

Sane Souvanhnakhoomman
De La Salle University-
Graduate program in Mechanical Engineering,
Manila, Philippine

*Abstract*: In today's agriculture, there are far too many innovations involved. One of the emerging technologies is pesticide spraying using drones. Manual pesticide spraying has a number of negative consequences for the people who are involved in the spraying operation. The result of exposure symptoms can include minor skin inflammation and birth abnormalities, tumors, genetic modifications, nerve and blood diseases, endocrinal interference, coma or death. However, Drone can be used to automate fertilizer application, pesticide spraying, and field tracking. This paper provides a concise overview of the use of drones for field inspection and pesticide spraying. displays different methodologies and controllers of agriculture drone and explains some essential Drone Hardware, Software elements and applications.

## 1. INTRODUCTION

The explosion of the human population makes high productivity, high performance and sustainable agriculture more important [1]. In the modern environment, agriculture is essential for the subsistence of more than 60 percent of total of the world's population. [2]. It is a critical element in the protection of the environment in the developed world. The agricultural modernisation is mandatory when demand and food supply are increased. Drone are one of the most advantageous equipment for modern agriculture.

The pesticides and fertilizers are critical components in the control of insects and the development of crops. Spraying pesticides and fertilizers by hand causes tumors, hypersensitivity, allergies, and other illnesses in people. Hence, Drone can be used to automate fertilizer application, pesticide spraying, and field tracking, which is also used for many applications such as search and rescue, police, code inspections, Emergency Management, fire. Other advantages of drones include their fast maneuverability, improved payload, high lifting power, and stability [3]. It comes with a universal sprayer for spraying both liquid and solid contents. The global nozzle sprays all pesticides and fertilizers, but the tension pump is only used when spraying pesticides and not when spraying fertilizer. In wide fields, the GPS can be used to automatically direct the quadcopter and power it remotely. A quad copter is piloted by an autopilot controller, and the payload is driven by an RF transmitter and motors. The figure 1 illustrates the pesticides spraying mechanism [4].

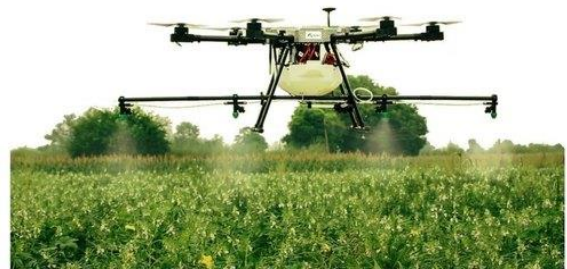

Figure 1: Pesticide spraying mechanism

This paper usually depicts the characteristics of appropriate Drones for a particular agricultural purpose. Furthermore, it will be clear as to which Drone archetype is needed for specific farming tasks. The systematic review of this article is based on basic keyword and abstract searches in Scopus, WOS (Web of Science), and Google Scholar databases. Several trustworthy websites were also consulted for subject-related material.

## 2. LITERATURE REVIEW

Yallappa (2017) improved an hexacopter with 6 BLDC motors and two LiPo batteries of 6 cells- 8000 mAh. Their research also includes a performance assessment of spray liquid discharge and pressure, spray liquid depletion, and droplet size and density determination. By means of their project, they eventually created a drone that can hold 5.5 L of liquid and has a 16-minute endurance period [5].

Dongyan (2015) investigated successful swath width and droplet distribution uniformity over aerial spraying systems such as the M-18B and Thrush 510G. The agricultural planes flowered respectively at 5m and 4m high, and by this experiment they conclude the disparity in swath width of M-18B and Thrush 510G in flight height [6].

Prof. B. Balaji (2018) created a hexacopter UAV for pesticide spraying as well as crop and environmental surveillance using Raspberry Pi and the Python programming language. Their UAV also has a variety of sensors, including DH11, LDR and

Kurkute (2018) used basic cost-efficient equipment to work with UAV quadcopter and its spraying system. Spraying with both liquid and solid material is done using the universal sprayer method. During their analysis, they also compared various agricultural controllers and came to the

Water Level Monitoring sensors. As a result of this experiment, they eventually concluded that with proper implementation of UAVs in the agricultural sector, savings in terms of water, chemical abuse, and labour can be projected to range between 20% and 90%. [7].

conclusion that the quadcopter system with the Atmega644PA is the most suitable due to its successful implementation [8].

Huang (2015) developed a low-volume sprayer that can be used in unmanned helicopters. The helicopter has a 3 m main rotor diameter and a payload capacity of 22.7 kg. It used to take at least a gallon of gas every 45 minutes. This research paved the way for the development of UAV aerial application systems for crop production with a higher goal rate and a larger VMD droplet scale [9].

Shilpa Kedari has suggested a low-cost, lightweight Quadcopter (QC) framework. The quadcopter is also known as Unmanned Aerial Vehicle (UAV). This quadcopter is compact and can be used for both indoor and outdoor crops. The quadcopter is an unmanned flight that uses an android smartphone to spray pesticides and fertilizer. The contact between the quadcopter and the android smartphone is achieved in real time using a Bluetooth device. This method is used to reduce agricultural field problems while still increasing agricultural yield [10].

Sadhana improved on the above approaches and created the quadcopter UAV and shower module, which can be used to spray pesticides in agriculture fields to increase efficiency and protect materials. The total load for this project is 1 kg and is used to spray low altitude pesticide quadruple copter lift. The Arduino UNO AT mega328 and Brushless Direct Current (BLDC), Electronic Speed Control (ESC), MPU-6050, which combines a MEMS accelerometer and a MEMS gyro into a single chip, Radio receiver, LIPO battery, and pesticide spraying module control this quadcopter [11].

### 3. METHODOLOGIES AND MATERIALS
#### *Methodologies*

The main board in the drone is the flight controller, which is loaded with cutting-edge firmware and is in charge of the actual flight. The flight controller controls a lot during the flight or drone at the same time. It has been designed and connected with a micro controller to the four motors without brush motor. BLDC motor attach in the Drone setup model with the rotors. These BLDC motors are controlled by the Electronic Speed controllers (ESC). The drone is powered by the transmitter and receiver of the radio network. There are several platforms for individual drone control activities for any RC transmitter. A sample block diagram shown in Fig 4. Different methodologies and controllers of Drone are shown in Table 3.

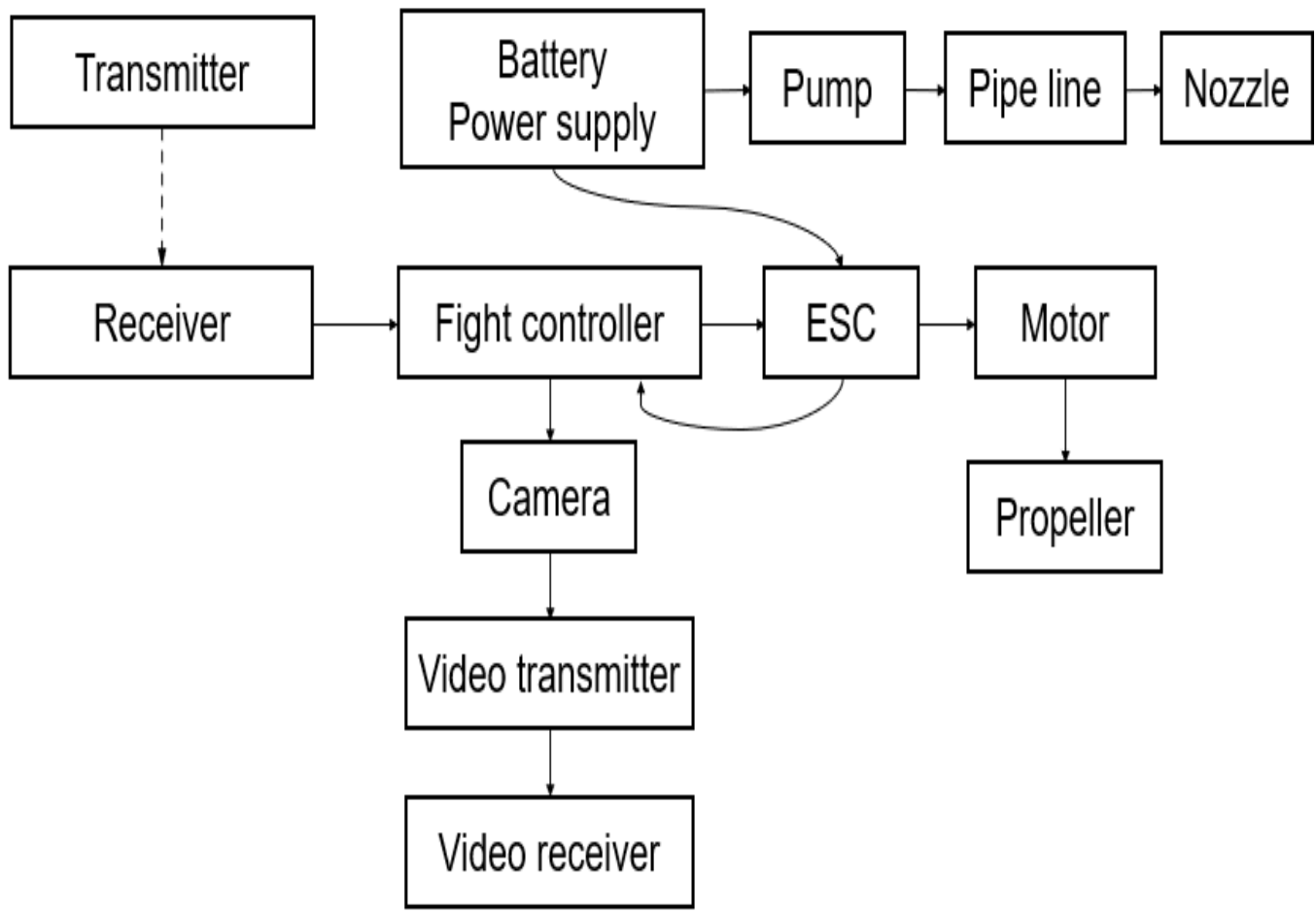

Figure 4. The main components of methodologies drone

Table 1. Different methodologies and controllers of Drone

| Author | Implementation Details | Components | Controller | Nozzle Type | Remarks | Load (L-Litres) |
|---|---|---|---|---|---|---|
| Munmun Ghosal (2018) [12] | Monitoring the exact place in which GPS module is notable for air pollution. | ESC, BLDC motor, sensor such as LM35, AM1001, LDR, MQ6, and MQ135. | Arduino Uno ATmega328 | | It is a low-cost, high-efficiency model. | |
| Sabikan (2016) [13] | A platform for the autonomous UAV quad copter was built for open-source projects. | IMU, 2.4GHz telemetry, ESC. | ArduPilotMega (APM) 2.6 | | The quad copter OSP offers both software and hardware as a comprehensive framework and also flexibility design for research or project purposes. | |
| Shilpa Kedari (2016) [10] | A quad copter is deployed on Android smartphone. These android applications control the quad-copter for pesticide and fertiliser spraying. | IMU, barometer, accelerometer, gyroscope. | Arduino board | | Reduces the problem of the health of farmers during pesticides and fertilizer application. | |
| Sadhana B (2017) [11] | Developed the quad copter and sprayer module | ESC, BLDC, MPU 6050 sensor. | Arduino Uno ATmega328 | Mini nozzle | High stability and increased power lifting. It is easy to compare the quad copter control to a miniature helicopter or vehicle. | 1kg |
| Parth N. Patel (2016) [14] | The quad copters enable the fabrication of unique folding frames for safe transport and convenient packaging of cylindrical cushioned boxes. | Accelerometer, gyroscope, IMU, Infrared camera, BLDC, ESC | Atmel AVR microcontroller | | It is adaptable, allows function performance to be modified and also allows technological integration. | |
| Weicai Qin (2019) [15] | Study the effect in different heights and sprayers of the spraying system. | GPS, digital temperature, humidity indicator, water sensitive. | N-3 type | Rotary atomizer | In this UAV the pesticides were initially employed in low altitude and low volume. | 25 lit |
| Tanga (2018) [16] | Determining the deposition of droplets in various forms. | Digital temperature, Humidity indicator, Water sensitive Sensors. Anemometer, Filter papers. | UAV ZHKU-0404-01 | Flat fan | For wind speed measurement. The indicator is used for air humidity measurement. | 15 L |
| Tejas S. Kabra (2017) [17] | Suggest Quad Copter [QC] to be introduced. The quad copter reduces the problem of farming | BLDC | | | This procedure reduces the medical problem created by hand sprinkling. | 1.5 to 3 L |
| Rahul Desale (2019) [18] | This project is being utilized by UAV in agriculture to spray insecticides. | BLDC, ESC, ratio controller, Transmitter. | Flight Controller | Fog nozzle | The benefit of this project is that it frame to spray pesticides in a safe place by utilizing drone. | |
| Shaik. Khamuruddeen (2019) [19] | This type is used for quad copter spraying of insecticides. | BLDC, ESC, Transceiver, Infrared Camera. | PID Micro Controller | | To identify less work where PSQ is used. | |

❑ **Hardware and software Components**

Drones are an array of sophisticated hardware, software and advanced technology. In general, many software and hardware components are used to properly control the drone according to Drone's startling variants. The unified body or structural components are usually referred to as hardware. Drone hardware shows the technical components of the Drone enough and takes software applications guidance. Furthermore, the software may be referred to as a Drone's brain. The software is supposed to tell the Drone whether or not to go and what to do. As a potential method of speculating and combining all of the Drone's critical data, a complicated structure benefits from the software portion. Drone software consists of a large number of different applications, processes, and operations. It also has a specific liability, as shown by its hardware components. The Drone is correctly controlled by a special combination of hardware and software. Tables 1 and 2 show some of the underlying hardware and software components respectively.

Table 2. The following are some of the most important drone hardware components and implementations.

| Name of the Element | Purpose | Reference |
| --- | --- | --- |
| BLDC motor | Movement control | [5][7][11][12][14] [17-19] |
| Flight Controller | control fixed-wing drone | [5-11][18] |
| Transmitter | For use radio signals to transmit commands wirelessly | [5][6][7][10][11][18][19] |
| ESC | Regulates the speed of BLDC | [5][7][9][11][12][13][14][19] |
| Propeller | Movement of drone | [11] |
| Water Pump: | for spraying water | [5][9][11] |
| GPS | Navigating | [5][6][7][9][11][15] |
| Camera | Record video or capture image | [5][9][11][14][19] |
| Accelerometer | For measure the acceleration Gyro | [5-11][12][14] |
| Gyroscopes | For rotational motion or Maintaining orientation and angular velocity | [5-11][12][14] |
| Magnetometer | Measuring the strength and direction of the magnetic field. | [5-11][12] |
| Battery | Retaining power | [5-11] |
| WSN | Sensing environmental conditions | [7][11] |

Table 3. The following are some of the most widely used Drone Software components and implementations.

| Name of the Element | Purpose | Reference |
| --- | --- | --- |
| C++ | Image processing | [12][13] |
| MATLAB | Image-processing and analysis | [20][21][22] |
| Adobe Photoshop | Distortion emendation | [23][24] |
| GIS | Capturing and analysing spatial and geographic data. | [25] [26] |
| MAVLink | Communicating with UAVs | [27][28] |
| Pix4D | Vegetation calculation and 3-D models construction | [29] [30] |
| Arduino | Control system | [10][11][12][13][14] |
| Python | Controlling | [7][31] |

## 4. CONCLUSION AND FUTURE WORK

The evaluation provided in Table 1 supports the use of Unmanaged Aerial Vehicle (UAV) in different quadcopters and improves the agricultural accuracy method the pesticides and fertilisers in agricultural fields in various crops. However, table 2 and 3 displayed some of the most important drone hardware components and implementations and the software is supposed to tell the Drone whether or not to go and what to do. As a potential method of speculating and combining all of the Drone's critical data. Drone software consists of a large number of different applications, processes, and operations

Drone is still in its early stage in precision agriculture and maybe a scope for additional improvement both in technology and in agriculture. It is expensed to develop Drone's innovation, improved ways of image processing, lower prices, flying times, battery, new camera models, small volume spraying systems and kinds of nozzles.